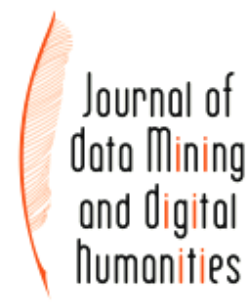

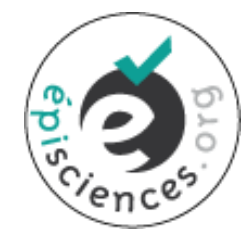

# Computer Analysis of Architecture Using Automatic Image Understanding

**Fan Wei[1], Yuan Li[2], Lior Shamir[3*]**

1 Lawrence Technological University, USA

*Corresponding author: Lior Shamir lshamir@mtu.edu

**Abstract**

In the past few years, computer vision and pattern recognition systems have been becoming increasingly more powerful, expanding the range of automatic tasks enabled by machine vision. Here we show that computer analysis of building images can perform quantitative analysis of architecture, and quantify similarities between city architectural styles in a quantitative fashion. Images of buildings from 18 cities and three countries were acquired using Google StreetView, and were used to train a machine vision system to automatically identify the location of the imaged building based on the image visual content. Experimental results show that the automatic computer analysis can automatically identify the geographical location of the StreetView image. More importantly, the algorithm was able to group the cities and countries and provide a phylogeny of the similarities between architectural styles as captured by StreetView images. These results demonstrate that computer vision and pattern recognition algorithms can perform the complex cognitive task of analyzing images of buildings, and can be used to measure and quantify visual similarities and differences between different styles of architectures. This experiment provides a new paradigm for studying architecture, based on a quantitative approach that can enhance the traditional manual observation and analysis. The source code used for the analysis is open and publicly available.

## INTRODUCTION

Architecture is one of the oldest forms of the combination of science and art. In addition to usability and environmental aspects of buildings, architecture has substantial aesthetic considerations, and the beauty of the building is considered one of the three most important aspects by which architecture is measured, along with the usability of the building and its durability (Vitruvius, 30BC). While beauty and aesthetics are subjective concepts that vary between different cultures (Carlson, 2002), architecture in different eras and geographic locations is fundamentally different, and the differences are the result of the complex combination of social, cultural, climatic, historical, religious, and geological influences (Fletcher, 1931). In addition to the subjections of beauty and its sensitivity to different cultures and societies (Saito, 2008), the differences between architectures is also the function of different available building materials and different building technologies (Devin & Nasar, 1989).

However, different cultures and societies are associated by influential links, making it difficult to consider any certain culture or society as an independent unit. Also, cultures can be divided into sub-divisions, related by era, geographical location, religion, etc. For instance, the Islamic culture can be divided into several sub divisions (Black, 2011), as the Islamic world spans over a very large territory, leading to different architectural styles in different regions and eras (Hillenbrand, 1994; Garlake, 1966; Petersen, 1996). The same is true for European architecture, featuring a complex profile impacted by the geographical location and era, reflecting social, religious, political and technological changes within the continent (Pevsner, 1972).

While these differences are often easy to notice by eye and can be characterized by basic architectural features (Devlin, 1990), the complex nature of art and architecture often makes these differences more difficult to quantify. For instance, given several architectural styles of interest, it is often difficult to determine which styles are more similar, and the decision needs to be made in a manner that often involves subjective analysis combined with identification of basic architectural features.

Previous work on automatic analysis of architecture focused on the automatic association of buildings with one of a crisp set of architectural styles (Shalunts et al., 2014; Xu et al., 2014; Zhang et al., 2014; Henn, 2012; Llamas et al., 2017; Guo & Li, 2017), or automatic classification of architectural elements (Bassier et al., 2017). Other work focused on systems that can use such techniques for the purpose of preservation and cultural heritage (Merchán et al., 2018). Another task of using computer analysis of visual architectural data is the ranking and estimation of urban environment quality (Liu et al, 2018), profiling how urban environments are perceived (Dubey et al., 2016), measuring the livelihood of neighborhoods (De Natai et al., 2016), or automatic estimation of a building age (Zeppelzauer et al., 2018; Lee et al., 2015).

Here we propose a computer method that can analyze architectural styles by applying computational image analysis of images of buildings. In contrast to some previously proposed computational methods for architecture classification (Shalunts et al., 2014; Xu et al., 2014; Zhang et al., 2014; Henn, 2012), the main goal of the method described in this paper is not to automatically associate a building with a certain architectural style, but to determine similarities between different architectural styles and provide a network of similarities between architectural styles of interest. The method provides a new paradigm for the studying of architecture history, as it uses objective quantitative approach that is not sensitive to the subjectivity of the viewer, and can identify influential links between architectures based on sets of images of the architecture.

## I DATA

The data used in this study are images of chosen cities and countries, all were collected using Google StreetView, which has been used in the past for automatic analysis of the changes in architectural style over time (Lee et al., 2015). The cities and countries that were selected for analysis are London, Paris, Brooklyn, Istanbul, Beijing, Bangkok, Buenos Aires, Johannesburg, Mexico City, Sydney, Wellington, Berlin, Frankfurt, Hamburg, Kiev, Madrid, Detroit, Stockholm, Russia, Korea, and Japan, representing diverse architectural styles. In addition to the dataset of 21 locations, a smaller dataset of 12 different locations was also used, and included: Bangkok, Berlin, Buenos Aires, Frankfurt, Hamburg, Johannesburg, Kiev, Madrid, Mexico City, Stockholm, Sydney, and Wellington.

The data were collected using Google StreetView by virtually travelling through streets in each of the selected cities. In the process of image collection, interference factors like sky, people, cars, and trees were avoided as much as possible. Therefore, views that had people or substantial portion of sky or vegetation were excluded from the database. Views that contained the shadows of other buildings or structures projected on them were also not used due to the impact of the shadow on the visual content. Also, any labels added to the image by Google StreetView (e.g., the address) were removed, so that the pattern recognition algorithm does not make use of these labels for making the analysis and classification. To make the dataset more consistent, the StreetView images were “typical” buildings in the target destination, and not iconic buildings or more modern office and residential buildings. While iconic architectural structures might be landmarks of their respective cities or countries, they might not reliably reflect the architecture of the city. For instance. The Eiffel tower is clearly the most famous architectural structure in Paris, but it does not reflect the architecture of the city (Aaseng, 2000). Therefore, the Google StreetView images of Paris were collected around the city center, but excluded structures such as the Eiffel tower, Arc de Triomphe, churches, or buildings in the financial district of La Defense.

For each destination, 50 images were used. That is, the only information used in the experiment for each destination is 50 images of buildings, and the analysis of the similarities between the architectural styles is determined by the similarities between the different sets of 50 images. Figure 1 shows example of a Google StreetView image used in the experiment. Figure 2 is a Google StreetView image that was not included in the experiment due to the shadow projected on the building and the plants.

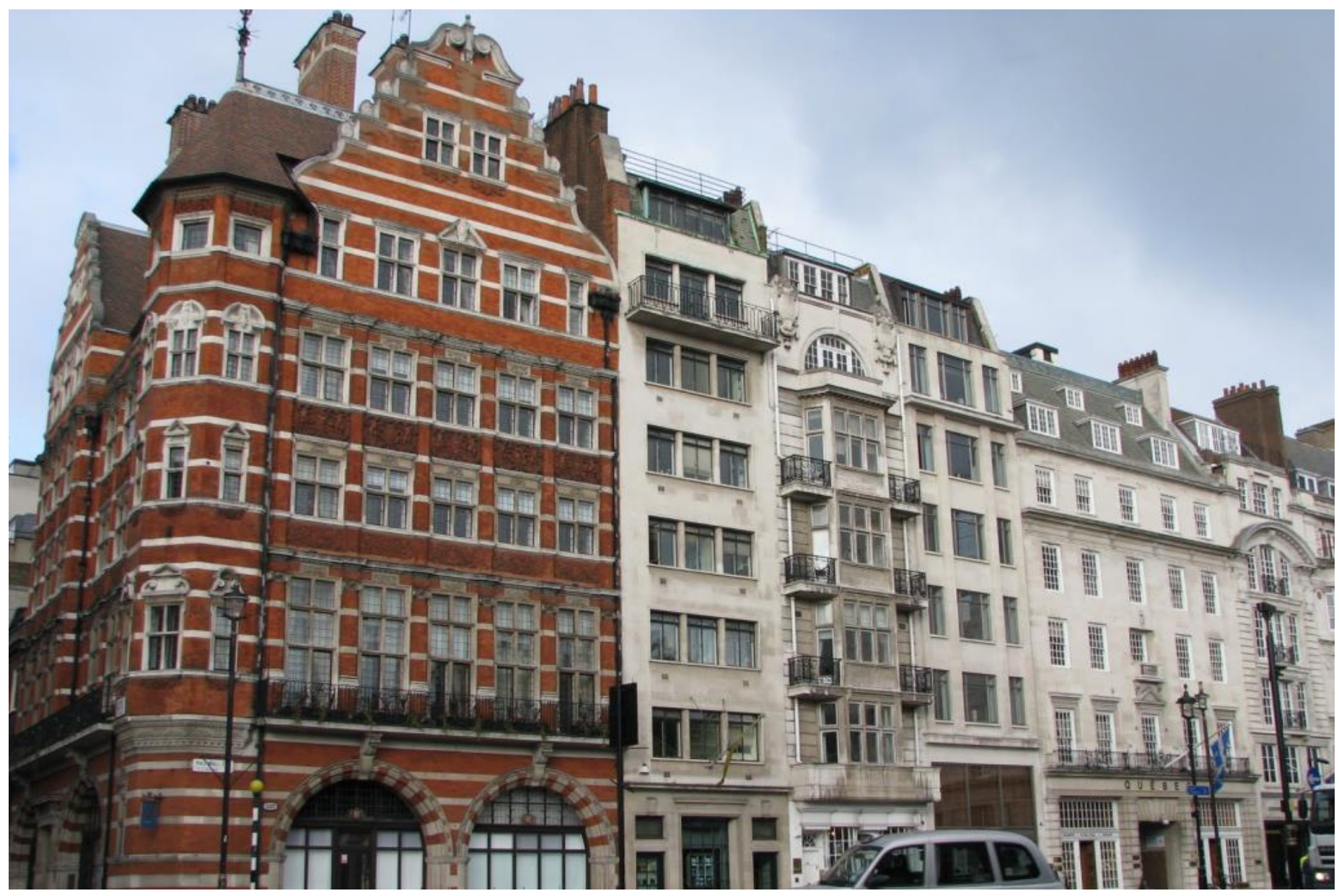

Figure 1. An example Google StreetView image used in the experiment. Labels added by Google StreetView such as the location, address, or logos were removed so that only natural data are contained in the image.

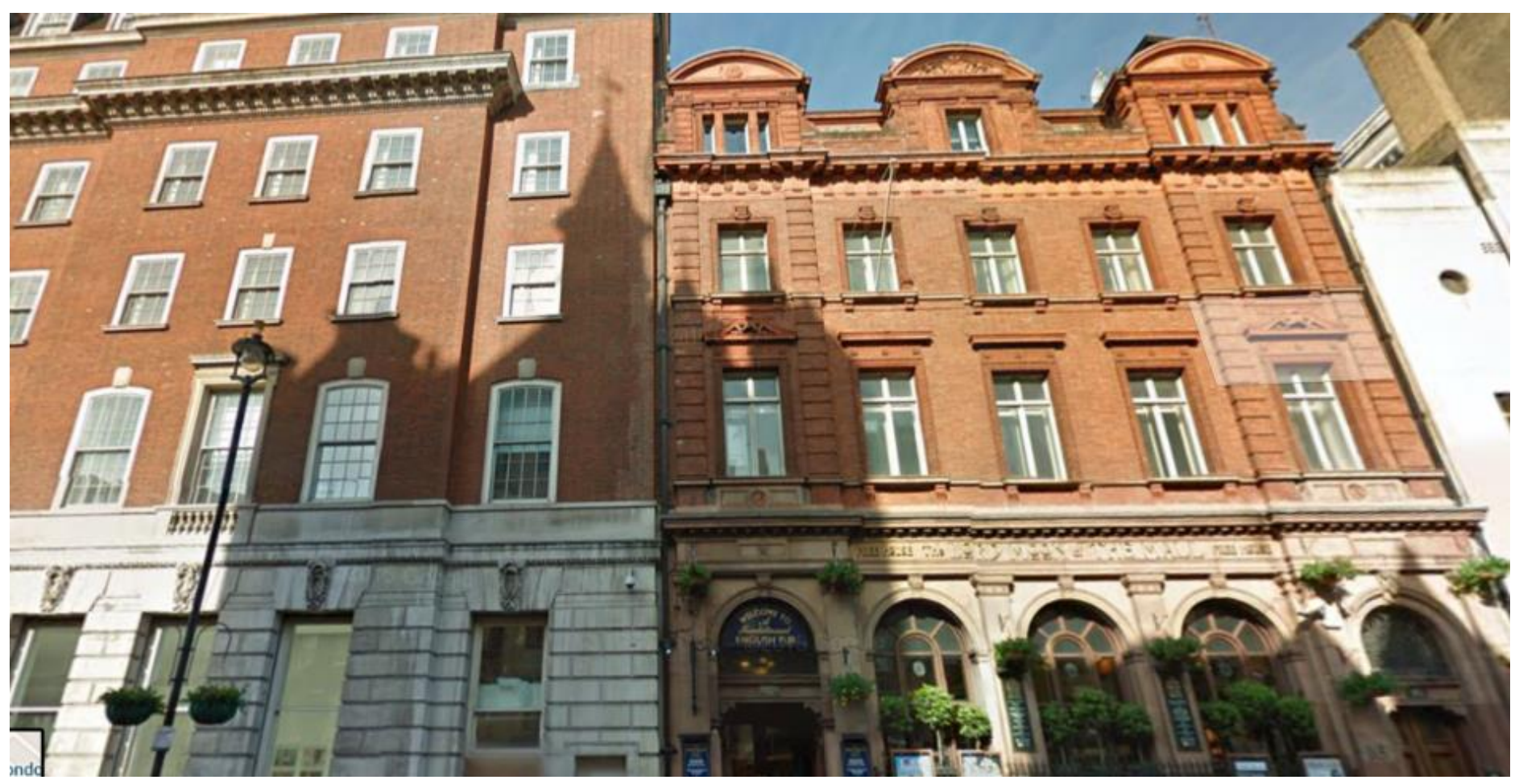

Figure 2. An example Google StreetView image that was excluded from the experiment due to the presence of shadow and plants.

The images were saved as Tagged Image File (TIF) format files, and then pre-processed by IrfanView (Skiljan , 2012) to automatically crop the images and separate the visual content from the labels added to them by Google StreetView. Since Google StreetView adds the labels at the consistent locations in all images, separating the natural visual content from the labels can be done in a batch. That resulted in a set of images with no labels or any other information other than the image of the building. All images were of dimensionality of 2000x1550 pixels.

## II IMAGE ANALYSIS

The image analysis is based on a data-driven supervised machine learning system. That is, the dataset is separated into training and test samples, and the computer is first trained from a set of images annotated with the name of the city where the image was taken, allowing the computer to identify patterns that are typical to the architecture of certain cities. Then, the efficacy of the system is tested by attempting to classify the test samples (that are not included in the training set), and measure the number of samples that were correctly associated by the computer with the places in which the images were taken. If the system is able to predict the location of the buildings and associate an image of a building with the correct city, the system can be considered informative and can be used to analyze architecture.

Due to the complex nature of associating an image of a building with the city it is part of, and the fact the some cities can share similar architectural styles that their similarity does not allow clear unique identification of each building, it is expected that an automatic system will not be able to associate a building with the correct city in all cases. However, if the system is able to predict the correct city in accuracy higher than the accuracy of mere chance guessing, it can be assumed that the system is informative, and it is sensitive to the architectural information inside the images.

The image analysis process is done by first converting all images into numerical values that describe the visual content. Then, the most informative features are selected, and the system automatically identifies patterns typical to specific architectural styles (training). These patterns are used to classify test sample and measuring the efficacy of the system by the correct recognition rate. A higher rate of associating the images with the correct city indicates that the system is more informative and more sensitive to architecture data, but as mentioned above, due to the complex nature of the problem perfect accuracy or close to that is not expected.

The images were analyzed using the Wndchrm image analysis scheme (Shamir et al., 2008). Wndchrm uses a comprehensive set of 4024 numerical image content descriptors that reflect shapes, colors, textures, fractals, and more, and has been used for various purposes ranging from astronomy (Shamir, 2009) to biomedicine (Shamir et al., 2009; Manning & Shamir, 2014). In particular, it has been used widely to study art history in a quantitative fashion by applying computational analysis to visual content (Shamir et al., 2010). For instance, it showed that the computer analysis of art is largely in agreement with how art historians view influential links between different schools of European art (Shamir & Tarakhovsky, 2012). It was also used to identify features typical to Jackson Pollock (Shamir, 2015) and show evidence of mathematical similarities between Jackson Pollock and Vincent von Gogh (Shamir, 2012). Another use of the Wndchrm scheme related to automatic analysis of art is the studying of art perception, showing patterns of differences between abstract expressionism and paintings by children and animals (Shamir et al., 2016). The numerical image content descriptors has been described thoroughly in previous papers, and the full technical description of these image features is available at (Shamir et al., 2008; Shamir et al., 2010; Shamir et al., 2009; Shamir & Tarakhovsky, 2012; Shamir et al., 2016; Burcoff & Shamir, 2017). Since Wndchrm has demonstrated its ability to analyze the complex visual content of paintings, it can be reasonably assumed that it can also be informative when analyzing visual content of architecture.

After the values of the numerical image content descriptors are computed for all images, the Fisher discriminant score (Bishop, 2006) of each feature is computed using the samples in the training set, and 85% of the features with the lowest scores are rejected, leaving the 604 most informative image features. Numerical image content descriptors with low Fisher discriminant are assumed uninformative for the purpose of analysis of visual data related to architecture, and are therefore rejected from the analysis.

Wndchrm analyzes the values of the numerical image content descriptors using the Weighted Nearest Neighbor (WND) algorithm (Shamir et al., 2008, 2009, 2010), such that the Fisher discriminant score assigned to each feature is used as its weight (Shamir et al., 2008, 2009, 2010). Then, the classification of each image is made by a Weighted Nearest Distance (Shamir et al., 2008, 2009, 2010) rule. Naturally, the predicted class of a given test sample is the class that has the lowest distance to that sample (Shamir et al., 2008, 2009, 2010).

In addition to the predicted class, the WND algorithm also provides a multidimensional distance between each pair of images in the database, as will be described in Section III. For each pair of architectural styles, the average distance between all pairs of images in these classes are averaged, and the averaged distances is used as a measure of similarity between these architectural styles (Shamir et al., 2008, 2014; Shamir & Tarakhovsky, 2012). These similarity values are then visualized by a phylogeny using the Phylip package (Felsenstein, 1993), providing a tree that visualizes the similarities between the architectural styles based on the collection of images of each location. Phylip was originally designed for visualizing similarities between organisms based on their genetics, but here the genetic information is replaced with similarities between images, so that Phylip visualizes similarities between sets of images representing architectural styles.

## III SEPARATING REGIONS OF INTEREST

The image analysis was done in two different manners. In the first, each image was separated into 16 equal-sized tiles, and each tile was treated as a separate image. That is, the numerical image content descriptors were computed for each tile separately, and each tile was classified separately. Then, the distance between image $I$ and class (architecture style) $A$ is measured by the average weighted distance between each of the 16 tiles and the tiles in class $A$ as shown by Equation 1

1) $D_{I,A} = \frac{\sum_{x \in I} d_{x,A}}{16}$,

where $d_{x,A}$ is the minimum weighted distance between the feature vector computed from tile $x$ of image I. Obviously, once an image is allocated to the training set, all of its tiles are also allocated to that set, to prevent a situation in which tiles from the same image can be present in both the training and test sets, allowing the algorithm to tiles that are part of the same image.

When all distances between the images of a certain class of architecture A and all other classes are computed, the similarity $M_{A,Q}$ between A and any of the other classes $Q$ is determined by Equation 2

2) $M_{A,Q} = \frac{\sum_{i \in Q} D_{i,A}}{|Q|}$.

Repeating that for all classes provides the distance matrix $M$, which is then normalized to the range of [0,1] by dividing the distance of each class to all other classes by the distance of the class to itself. The distance matrix is transformed into a similarity matrix by simply subtracting the distances from 1. As described in Section II, the similarity matrix can be visualized using

Phylip, with randomize input order of sequences where 97 is the seed, 10 jumbles, and Equal-Daylight (Felsenstein, 2002) arc optimization. That produces a tree of similarities that visualizes the similarity matrix.

In addition to the separation of each image into 16 equal-sized tiles, the experiment was also done by first detecting 16 small 100x100 regions of interest (ROIs) from each image. Once the regions of interest are detected, the 16 regions of interest of each image are used in the same manner the 16 tiles are used as described earlier in this section.

The regions of interest are detected by scanning each image with a 100x100 shifted window, and computing the standard deviation of the pixel intensities in each position of shifted window. Then, the standard deviations are sorted, and the 16 windows with the highest standard deviation are selected as the regions of interest. If a selected region of interest overlaps with a region of interest with a higher standard deviation, that region of interest is excluded to ensure that the regions of interest contain different parts of the image.

The intuition of selecting regions of interest by the standard deviation of pixel intensities is that sky parts of the image or flat walls do not contain important information that allows differentiating one architectural style from another. These areas will have lower standard deviation compared to areas that contain more visual features. Figure 3 shows an example of an image, and Figure 4 shows the regions of interests that were separated from it by using the standard deviation of the pixel intensities. As the figures show, the regions of interest that were separated from the image based on the standard deviation do not contain parts of the sky or flat floors or walls.

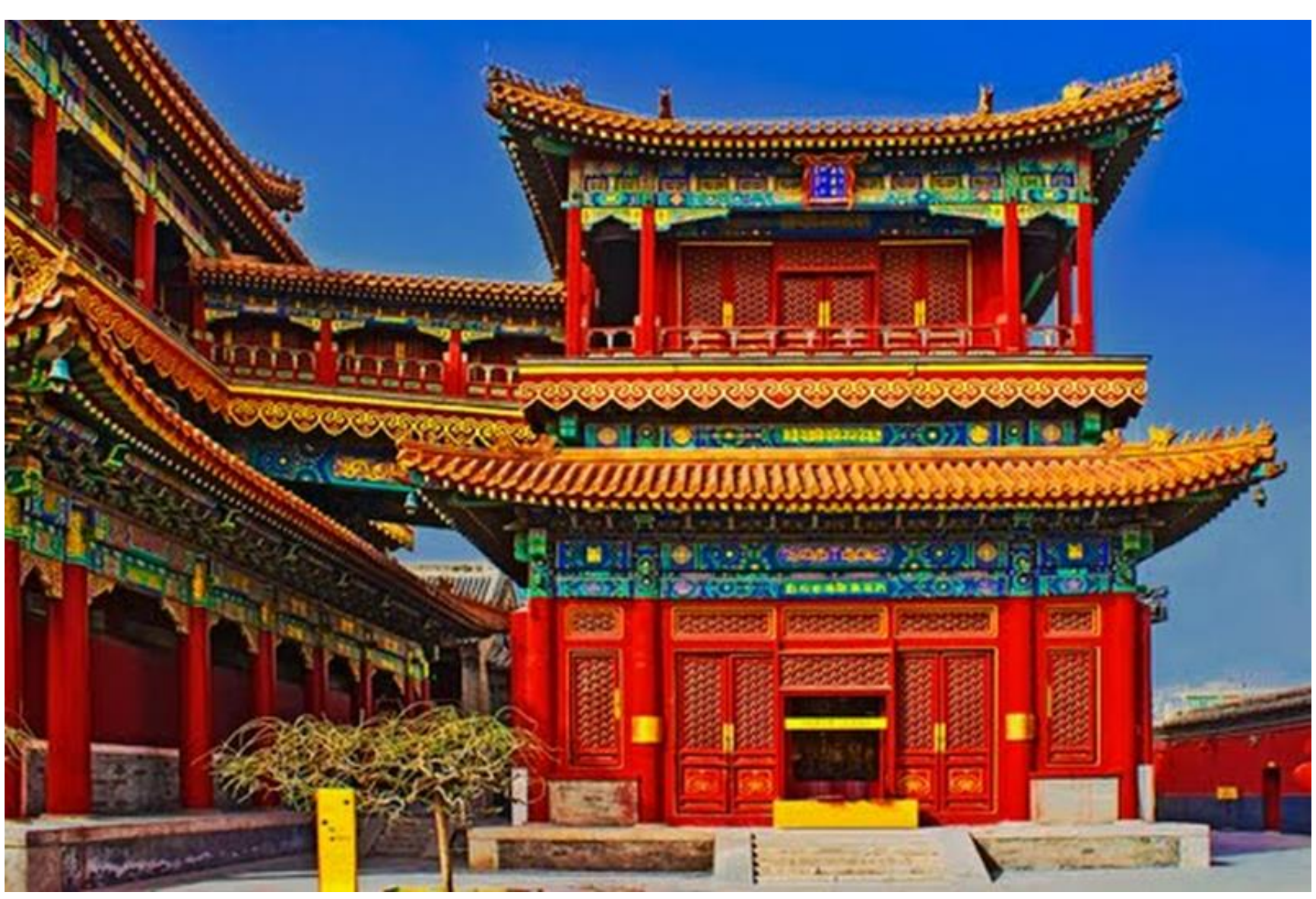

Figure 3. An example of the original image

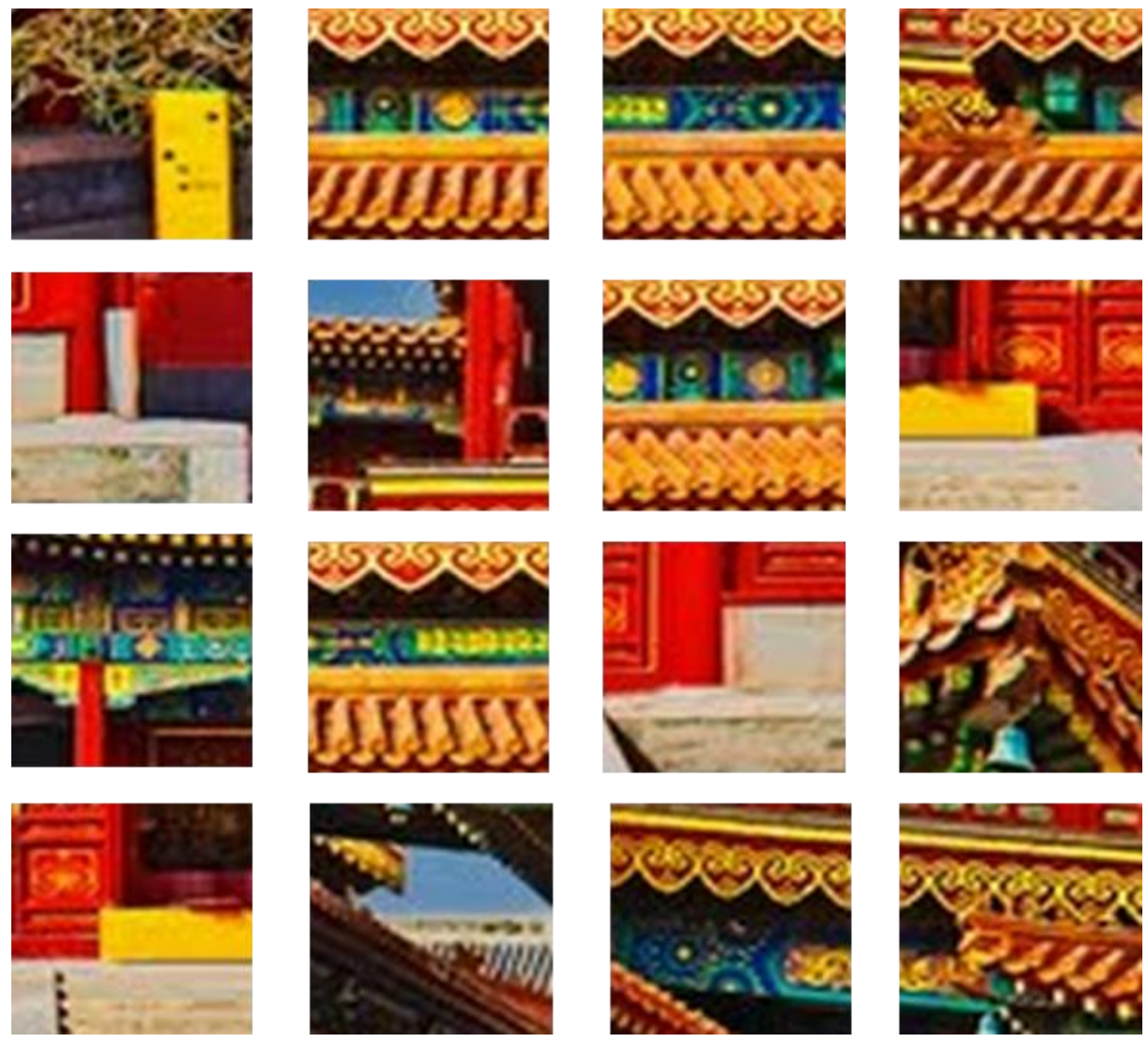

Figure 4. Top 16 100x100 regions of interest separated from the image. The regions of interest do not contain sky areas or flat walls and floors.

## IV RESULTS

The experiments were performed such that 45 images from each class were used for training, and five for testing. Each experiment was repeated 20 times such that in each run different images were randomly allocated for training and test sets. To first identify whether the algorithm is sensitive to architecture data, the accuracy of which a building image can be associated with a city was tested. The classification accuracy was measured by the number of correct classifications of the city based on the image of the building, divided by the total number of classification attempts. The classification accuracy when the original images were used was ~34%, and was elevated to ~41% when just regions of interest were used in the analysis. These numbers are much higher than mere chance accuracy, which is less than 5%. The fact that the system can associate an image of a building with a city based on images of other buildings in the same city shows that the algorithm is capable of identifying the patterns that are typical to the buildings in the different cities.

When the smaller dataset was used the classification accuracy was ~44% and ~59% for the original and regions of interest, respectivly. These numbers are also much higher than the accuracy of mere chance classification, which is ~8%. These numbers provide evidence that the computer algorithm is able to identify the architectural style based on the images, showing that the algorithm is sensitive to the architetural information contained in them. When using regions of interest separated from the images, the classification accuracy increases. The reason for that can be that when just regions of interest are used, the images contain less parts such as sky or flat walls, which might not contain information that can differentiate between architectural styles. These areas do not contain useful information, and might add noise to the learning process.

As described in Section II, the purpose of the algorithm is not merely to associate an image of a building with the place in which it was taken, but also provides a network of similarities between different target locations based on the similarities between the buildings and architectural styles. The network of similarities between each pair of cities is visualized using a phylogeny. Figures 4 and 5 show automatic phylogenies generated by the algorithm using just the Google StreetView images taken at each location. Figure 4 uses the entire images, while Figure 5 uses the regions of interest separated from each image.

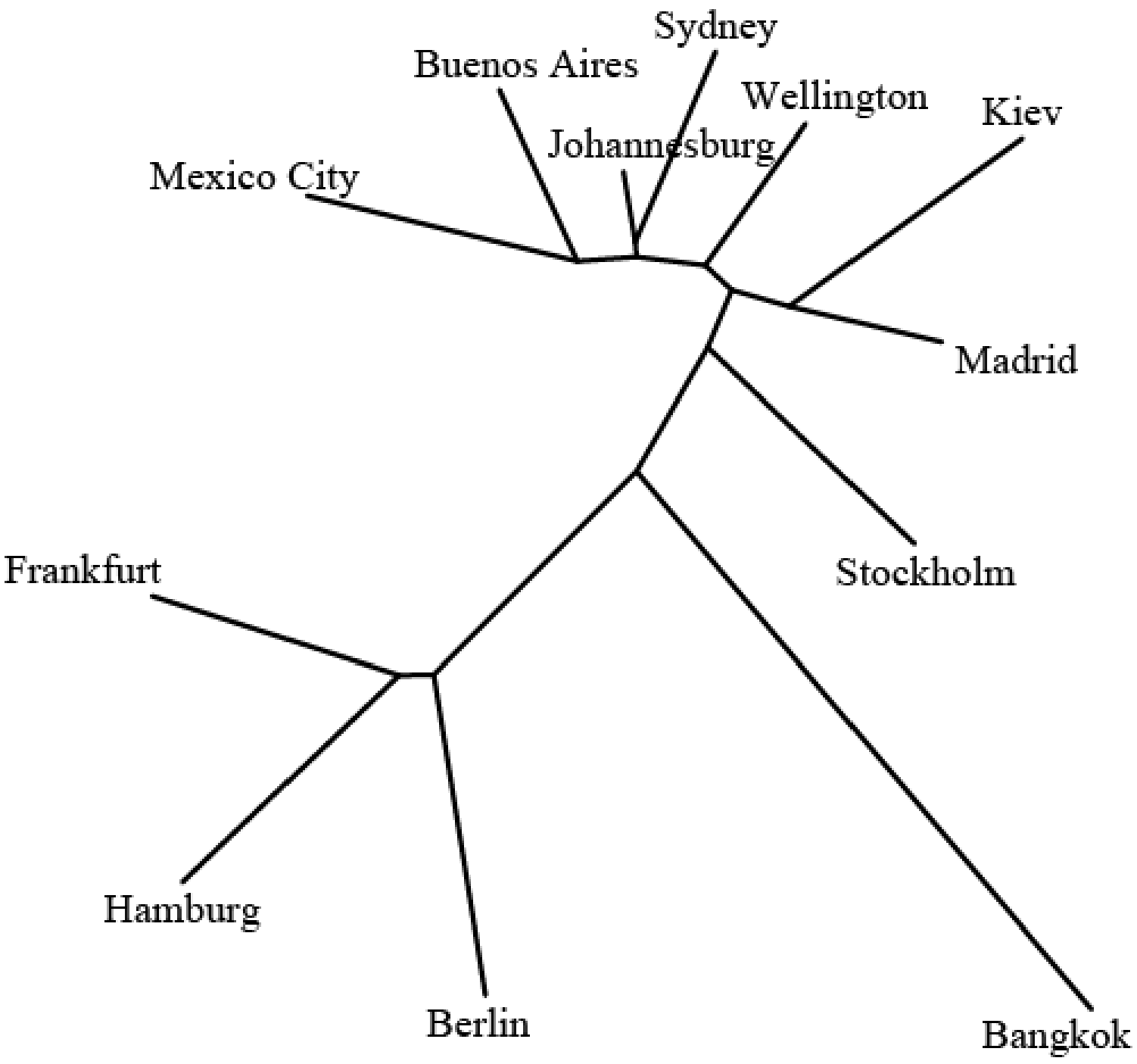

Figure 4. Phylogeny generated automatically by the information in Google StreetView images of buildings in different cities. The three German cities are clustered close to each other. The European cities are in the right side of the phylogeny. Bangkok, the only South Asian city in the dataset, is placed far from all the other cities.

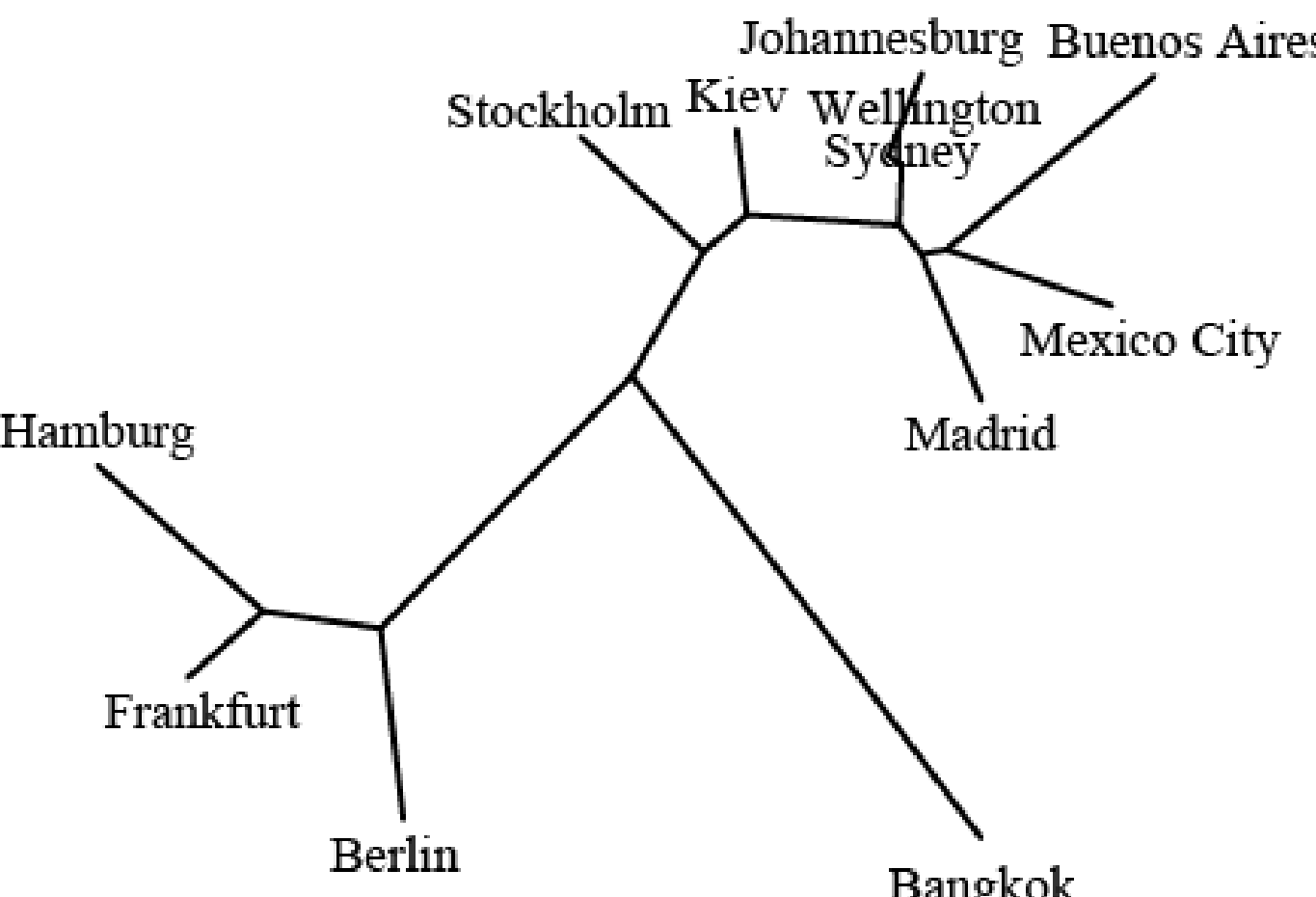

Figure 5. Phylogeny created automatically by the computer based on the analysis of 16 regions of interest separated from each of the 50 images of each location. The former Spanish colonies are positioned close to each other, while the former British colonies are also clustered together, suggesting similarities between the architectural styles identified by the computer.

In the phylogeny, the length of the edges reflects the similarity between the architecture of the two nodes, such that a shorter edge means that the architecture of the two nodes is more similar compared to other pairs of nodes connected with longer edges. The similarity between nodes that do not have a single edge connecting them is reflected by the length of the path between them, such that a shorter path indicates higher similarity between the two nodes.

As both figures show, the phylogeny that was automatically generated by the computer is largely in agreement with the geographic locations. The phylogeny that was generated when using the regions of interest seems better aligned with the cultural links between the different locations. For instance, the three German cities Hamburg, Frankfurt, and Berlin are positioned close to each other. Bangkok is clearly different from all the other architectural styles, and is indeed positioned far from all the other cities. Figure 5 shows that the former Spanish colonies New Mexico and Buenos Aires are positioned close to each other (Carson et al., 1981), and also close to Madrid, indicating that some architectural similarities are identified between Spanish architecture and the architecture in the colonies. Similarly, Johannesburg, Sidney and Wellington are clustered together. Although the three cities are geographically distant from each other, they are all former British colonies, and therefore it is possible that the architectural styles also share common characteristics of the British colonial architecture (Home, 2013). British colonies architecture can also be similar not just for the shared cultural and social impact of Great Britain, but also by the planning laws enforced by it in the colonies (Home, 1993). Therefore, while the phylogeny is not necessarily organized by the geographic locations of the cities, it is organized by the social and cultural influences. When using the regions of interest, the phylogeny is in better agreement with the cultural and historical links between the cities.

The larger dataset includes 21 different locations, each is represented in the database by 50 images taken using Google StreetView. Figure 6 displays the phylogeny that was generated automatically by the computer when using the original StreetView images, and Figure 7 displays the phylogeny generated by the computer when using the regions of interest from each image.

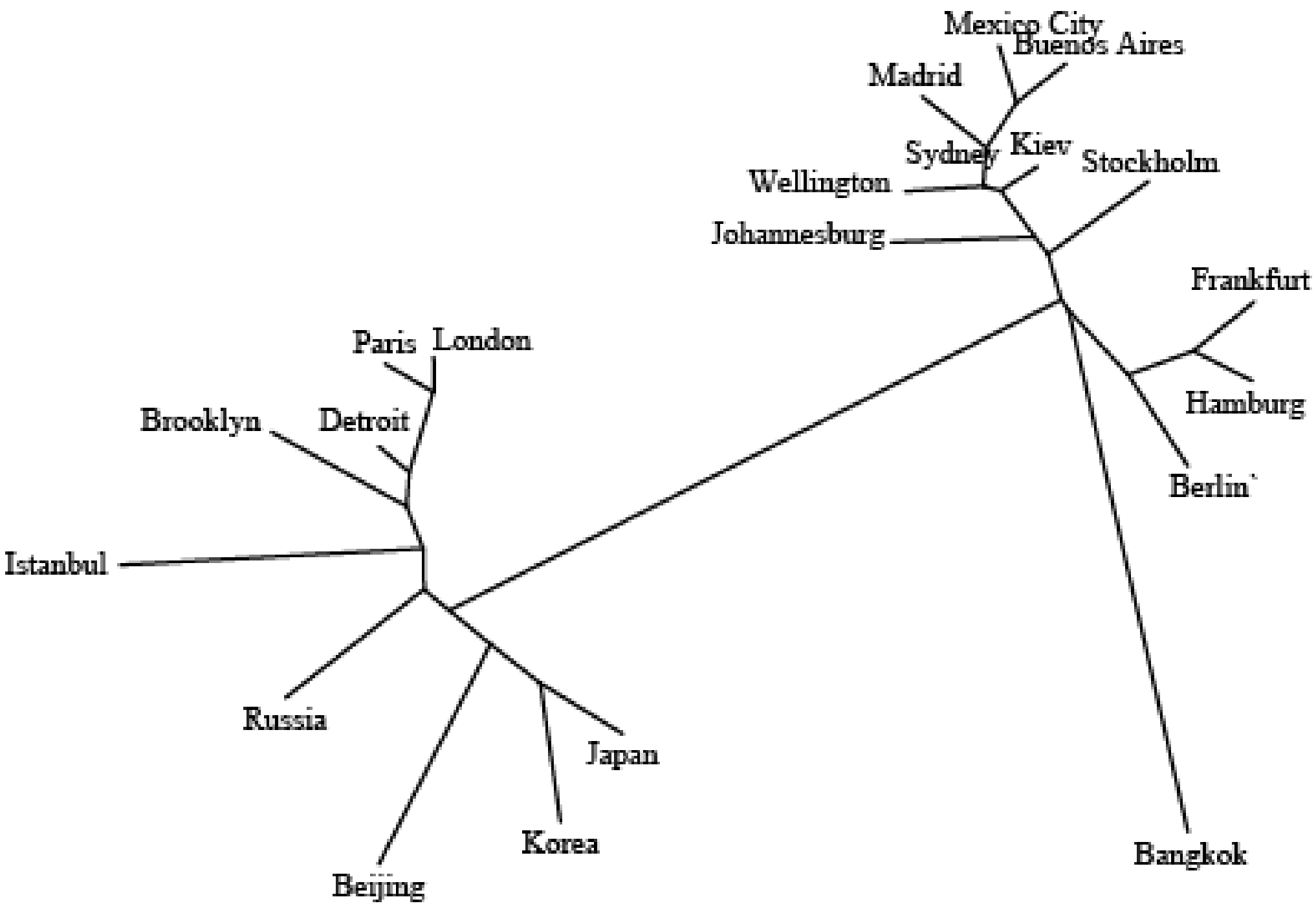

Figure 6. The computer generated phylogeny of 21 cities and countries. The phylogeny shows that the computer positioned the East Asian cities and countries close to each other, as well as the German cities.

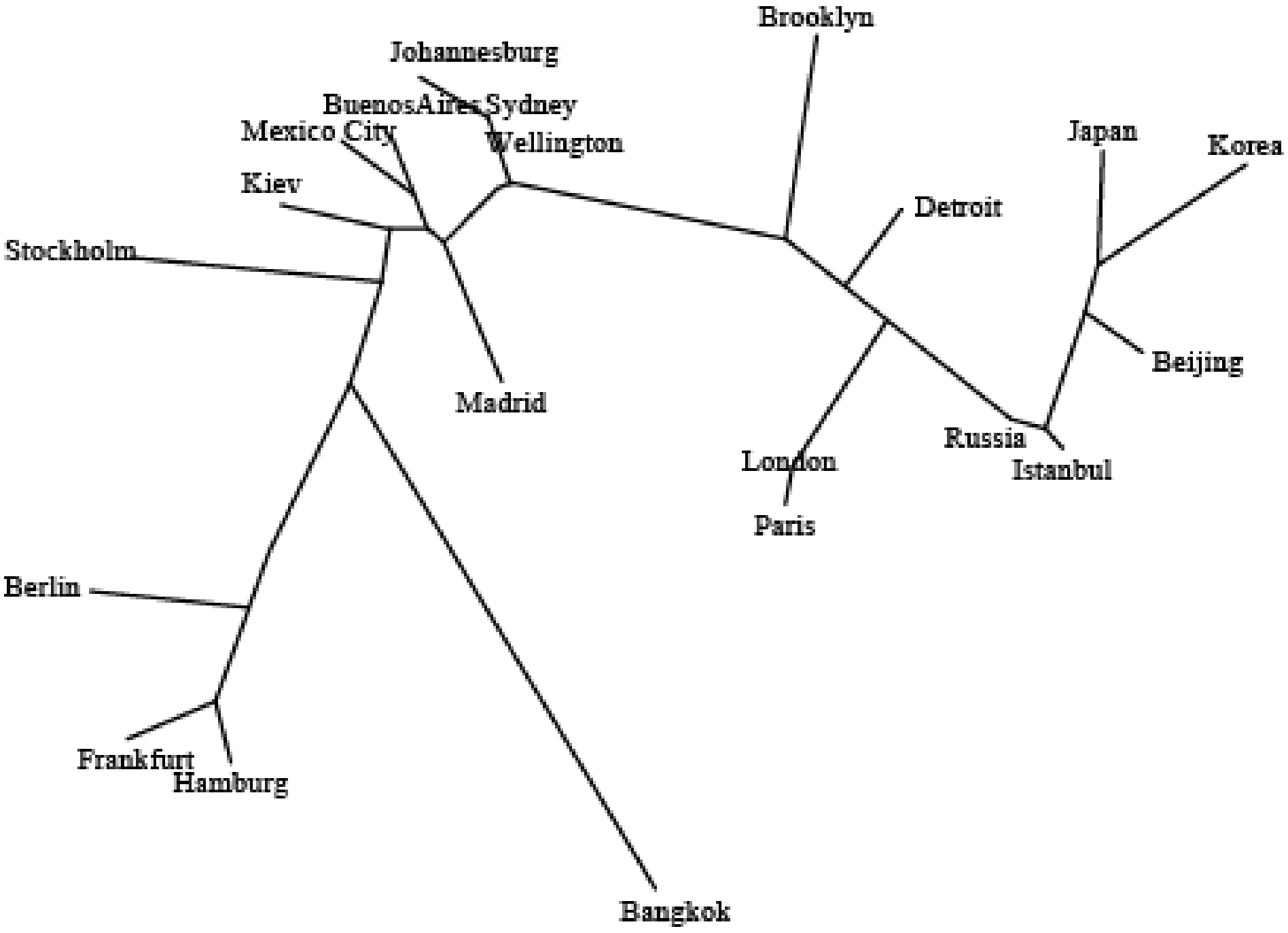

Figure 7. The computer generated phylogeny when using the regions of interest separated from the images. The three German cities Berlin, Frankfurt, and Hamburg are clustered together, as well as the former British colonies Johannesburg, Sidney, and Wellington. On the other side of the phylogeny the computer positioned together the East Asian locations. Former Spanish colonies Buenos Aires and Mexico City were positioned by the computer close to each other.

Comparing the figures using regions of interests separated from the images provided results that are in better agreement with the social and cultural links of the different locations. As Figure 7 shows, Berlin, Frankfurt and Hamburg were all positioned close to each other, and the East Asian locations Korea, Japan, and Beijing were also placed by the computer close to each other. The former British colonies Wellington, Sydney, and Johannesburg are positioned near each other, as well as the former Spanish colonies New Mexico and Buenos Aires. South Asian Bangkok is placed far from all the other cities.

These results show that the computer is able to reconstruct links of similarities between architectural styles in a fashion that largely agrees with the cultural links based solely on analyzing the images of the architecture, and without using any metadata or other information that is not in the image. The agreement between the similarities identified by the computer and the cultural links between the different locations show that the computer analysis is sensitive to the architectural styles as reflected by the images of the buildings. When using regions of interest the computer can analyze the architecture automatically in a fashion that better reflects the architectural styles.

## V CONCLUSIONS

Architecture is influenced by a combination of social, cultural, climatic, historical, religious, and geological aspects that shape architectural styles (Fletcher, 1931). Since none of these elements can be isolated, it is clear that different architectural styles have strong influential links, and some architectural styles are more similar to each other compared to other styles (Devlin, 1990).

Since the analysis of architecture is highly complex, the comparison of different architectural styles is a non-trivial task, requiring close manual observations of the features that characterize the architecture. Therefore, the analysis is subjective and involves the perception of the person examining the architecture.

As urban sensing systems have been becoming increasingly more prevalent (Lane et al., 2008), urban data is becoming easily accessible, and can be used for quantitative data-driven research that would have been very difficult to perform in the pre-information era. Here we show that machine vision can analyze images of buildings and provide information about their architectural styles in the context of other building images representing different schools of architecture. Based on that method we propose a quantitative method to study architectural styles using computer analysis. The method is not dependent on the human perception of architecture, and provides a network of similarities in the form of a phylogeny, based solely on images of buildings representing the architecture. That form of analysis is a new tool in architecture history research, allowing the studying of the history of architecture in a fully quantitative and automatic manner, and without human intervention.

In this study, just residential buildings are used, but the method can be used to analyze different forms of buildings and other structures. For instance, in this study building that serve a religious purpose were excluded from the research, but as architecture is tightly related to religion, it is clear that similar methodology could be used to study that link.

The source code for the Wndchrm method used in the analysis is freely available at http://www.ksu.edu/lshamir/downloads/ImageClassifier/